%% file: acl_latex.tex
\pdfoutput=1

\documentclass[11pt]{article}

\usepackage[final]{acl}

\usepackage{times}
\usepackage{latexsym}

\usepackage[T1]{fontenc}

\usepackage[utf8]{inputenc}
\usepackage{amsmath}
\usepackage{microtype}

\usepackage{inconsolata}

\usepackage{graphicx}

\title{RELexED: Retrieval-Enhanced Legal Summarization with \\ Exemplar Diversity}

\author{Santosh T.Y.S.S, Chen Jia, Patrick Goroncy, Matthias Grabmair \\ School of Computation, Information, and Technology; \\
Technical University of Munich, Germany \\ \ }

\begin{document}
\maketitle
\begin{abstract}
This paper addresses the task of legal summarization, which involves distilling complex legal documents into concise, coherent summaries. Current approaches often struggle with content theme deviation and inconsistent writing styles due to their reliance solely on source documents. We propose RELexED, a retrieval-augmented framework that utilizes exemplar summaries along with the source document to guide the model. RELexED employs a two-stage exemplar selection strategy, leveraging a determinantal point process to balance the trade-off between similarity of exemplars to the query and diversity among exemplars, with scores computed via influence functions.  Experimental results on two legal summarization datasets demonstrate that RELexED significantly outperforms models that do not utilize exemplars and those that rely solely on similarity-based exemplar selection.
\end{abstract}

\section{Introduction}
\input{text/introduction}

\section{RELexED: Our Method}
\input{text/method}

\section{Experiments}
\input{text/experiments}

\section{Conclusion}
\input{text/conclusion}

\section*{Limitations}
\input{text/limitations}

\section*{Ethics Statement}
\input{text/ethics}

\bibliography{custom}

\appendix

\input{text/appendix}

\end{document}

%% file: text/introduction.tex
Legal summarization is an essential task, aimed at distilling complex legal documents, such as court rulings and case judgments, into concise summaries \cite{farzindar2004atefeh,saravanan2006improving}. These summaries assist legal professionals in quickly grasping the core information, saving time and enhancing decision-making efficiency \cite{grover2003automatic,grover2003summarising}. However, summarizing legal texts is uniquely challenging due to the highly structured and formalized nature of legal writing. Legal summaries must maintain factual accuracy, adhere to specific terminologies, and follow established patterns that reflect the reasoning and argumentation typical in legal documents \cite{bhattacharya2019comparative,bhattacharya2021incorporating,deroy2023ready}.

Current approaches to legal summarization often rely solely on the content of the source document, leading to problems such as content deviation, where generated summaries stray from the central themes of the case and inconsistent writing style, failing to match the rigor and formality required in legal contexts \cite{shukla2022legal,moro2022semantic,elaraby2022arglegalsumm,shen2022multi,santosh2024beyond,santosh2024lexsumm}. This occurs because existing models are expected to implicitly learn the templatization process during training, which is prevalent in legal writing where they follow specific templates or writing structures. To address this, it is essential to introduce writing templates as exemplars that can bring not only guidance of writing format but also additional background knowledge necessary \cite{oya2014template,gao2019write}. Rather than manually creating templates for each scenario which turns expensive to obtain from domain experts, one scalable way is to use reference summaries in the training corpus as exemplars \cite{an2021retrievalsum,wang2022training}. 

To the best of our knowledge, there has been no exploration of retrieval-enhanced exemplars for legal summarization task. This approach consists of two modules: Retriever and Summarizer. Given a query document, the Retriever is asked to retrieve the most related exemplars from the training corpus. The second module Summarizer generates the summary not only relying on source document but also the retrieved exemplars. This method bears resemblance to recent works of in-context learning \cite{brown2020language,dong2022survey} which appends a few demonstration examples to the query to form a prompt which is then fed into the language model for prediction. In this work, we are mainly interested in supervised learning where the model parameters are fine-tuned to learn from given examples than in-context learning which does not perform parameter updates and expects the model to learn the pattern using the demonstration examples in the prompt. The success of retrieval-enhanced methods in supervised learning demonstrates that, even with vast numbers of parameters, models are unable to memorize every pattern from the training data and hence by retrieving relevant examples and explicitly providing them to the model can enhance its performance \cite{wang2022training,tyss2024mind} . This approach suggests that rather than constantly scaling models to larger sizes, we can achieve high-quality results by equipping a moderately sized model with training data that closely matches the current instance, thereby not only boosts performance but also significantly reduces computational costs \cite{an2021retrievalsum}. 

The main crucial link of retrieval-enhanced summarization are the chosen exemplars. Previous approaches generally rely on retrieving examples based on the similarity between the input and the examples based on traditional lexical-based BM25 \cite{cao2018retrieve,wang2022training} or embeddings \cite{an2021retrievalsum}. Despite the improved performance, we posit that similarity based exemplar selection do not always take into account the inter-relationship between different exemplars. For instance, the ignorance of redundancy among examples can result in almost identical examples, providing no additional information to the model.

In this paper, we propose a two-stage exemplar selection strategy for legal summarization, RELexED which balances both the quality and the diversity among the selected exemplars. In the first stage, we rely on semantic similarity to filter out unrelated examples and then in the second stage, instead of selecting each example independently, RELexED captures the inter-relationship between selected examples. To model the joint probability of selected set of examples given a specific input, we leverage  determinantal point process (DPP) \cite{kulesza2012determinantal} that learns to select relevant yet the most diverse example set. To measure the quality or similarity criterion, we resort to recent instance-based explanation methods \cite{koh2017understanding,pruthi2020estimating} which aim to quantify how a training example affects the prediction of a test example after training. Our experiments on two legal summarization datasets, SuperSCOTUS and CivilSum demonstrate the effectiveness of retrieval-enhanced summarization and specifically, our proposed exemplar selection method, RELexED which goes beyond using similarity-based selection only.

%% file: text/method.tex
We first introduce the framework for retrieval-enhanced summarization and the background of the Determinantal Point Process (DPP). Then, we introduce the our two-stage exemplar selection strategy, RELexED, balancing quality and diversity.

\subsection{Preliminaries}
\noindent \textbf{Retrieval-enhanced Summarization} First, we index the training corpus of document-summary pairs into a list of key-value pairs where the document serves as the key and its summary as the value. Given the input x, the retriever returns the $k$ exemplars from the training corpus. Then, these retrieved results are combined with the input $x$ to feed into the summarization model to generate the summary. Given that legal documents tend to be lengthy, we only incorporate the values (summaries) from the exemplars alongside the query input to the summarization model.

\noindent \textbf{DPP} DPP is a probabilistic model to express interactions between items and could helps us to select a representative subset while keeping high diversity among different items\cite{kulesza2012determinantal,chen2018fast}. Formally, let $S = \{1,\ldots, |S|\}$ denote a finite set of items and DPP defines a probability distribution over an exponential number of sets (all $2^{|S|}$ subsets of $S$), parameterized by a single $|S|\times|S|$ positive semi-definite kernel matrix, denoted as $L$. If $k$ is a random set of elements drawn from $S$, then the probability of selecting that subset is given by determinants of sub-matrix of $L$:
\begin{equation}
\scalebox{0.9}{$
p(k ; L) = \frac{\text{det}(L_k)}{\text{det}(L+I)} \quad  \sum_{k}\text{det}(L_k) = \text{det}(L + I)$}
\end{equation}
Here, $\text{det}(.)$ is the determinant of a matrix, $I$ is the identity matrix and $L_k$ is a sub-matrix of $L$ containing only entries indexed by elements of $k \subseteq S$. \citet{kulesza2012determinantal} provide a decomposition of the $L$-ensemble matrix allowing the modeling of relevance and dissimilarity independently and combining them into a single unified formulation with $L_{ij} = q_i\!\cdot s_{ij}\!\cdot q_j$, where $q_i$ is a positive real number indicating the quality/relevance of item $i$, and $s_{ij}$ captures the similarity between items $i$ and $j$. To understand why DPP serve as a trade-off between quality and diversity, please refer to the detailed explanation in App. \ref{dpp_exp}. The MAP inference for DPP involves sub-modular maximization, which is NP-hard. Therefore we use greedy algorithm for faster inference \cite{chen2018fast}. It begins with an empty set and iteratively adds each item to the selected set $Y$. The chosen item $i$ in each iteration is the one that maximizes the determinant value when added to the current selected set.
\begin{equation*}
\resizebox{0.7\linewidth}{!}{$
\begin{aligned}
    c &= \arg\max_{i \in S-Y } \left[ f(Y \cup \{i\}) - f(Y) \right] \\
    & \text{where} \quad f(Y) = \log \det(L_Y) 
\end{aligned}
$}
\end{equation*}

\subsection{Two-stage Exemplar Selection}
In the first stage, we leverage semantic similarity and utilize the BM25 to retreive the top \(k_1\) documents similar to the query from the training corpus. This step helps to eliminate irrelevant training instances that is not related to the given input and also aids to reduce the size of the candidate set, which simplifies the subsequent computations of the DPP matrix elements. The second stage involves employing DPP to select the top \(k\) exemplars from these \(k_1\), balancing relevance and diversity. The core question that arises in this stage is how to compute the quality/relevance score for each individual item with respect to the query item and the similarity score for pairs of items.

We leverage influence functions, which aims to trace a model’s predictions back to the most responsible training examples \cite{koh2017understanding}. For a model \( f \) with parameters \( \theta \) and loss function \( l(f_{\theta}, \cdot) \), the gradient \( g(\theta, \cdot) \) for a sample \( z \) is given by  \( g(f_{\theta}, z) = \nabla l(f_{\theta}, z).\) \citet{pruthi2020estimating} provides TracIn which is a gradient-only alternative to influence function approximation and the influence of a sample \( z \) on sample \( z' \) is given by
\[
\text{TracIn}(f_{\theta}, z, z') = g(f_{\theta}, z) \cdot g(f_{\theta}, z')
\]
\noindent We use an auxiliary model (similar to summarization model) fine-tuned to generate a summary directly from the input alone without exemplars. Given computational expensive nature to compute these, we follow \citet{thakkar2023self} to leverage the layer agnostic nature of TracIn and use only first layer of the encoder to compute the influence score. For quality/relevance score $q_i$ we compute the influence score of item $i$ on the query item $x$. For similarity score $s_{ij}$, we compute the influence score of item $i$ on item $j$. Once we populate the kernel matrix $L$, we greedily select items iteratively till the pre-defined number of exemplars that fit into the summarization model and concatenate them in the same order of selection.

%% file: text/experiments.tex
\subsection{Datasets \& Metrics}
We experiment with two legal summarization datasets. \textbf{SuperSCOTUS} \cite{fang2023super} is a multi-source dataset of U.S. Supreme Court (SCOTUS) cases. We use the summarization dataset provided, which uses the majority opinion of the court as the input to generate the Syllabus, which provides a summary of case. It consists of 4058 case-summary pairs, split into 3246/406/406 for training, validation, and test  with an average length of document and summary as 5405.46 and 902.23 tokens respectively. \textbf{CivilSum} \cite{malik2024civilsum} contains 23350 court decisions from the Supreme Court  and High Courts of India paired with human-written summaries. It is split into 21015/1168/1167 for training, validation and test. It has an average document and summary length of 2639.49 and 131.99 tokens respectively. 

We evaluate the quality of the generated summaries using ROUGE-1,2,L \cite{lin2004rouge} for lexical overlap with the reference paragraph, BERTScore \cite{zhang2019bertscore} for semantic similarity beween generated and referecne summary. For faithfulness, we report AlignScore \cite{zha2023alignscore} for factual consistency based on a unified alignment function between the input context and generated text. We also report coherence and fluency scores using UniEval metric \cite{zhong2022towards} based on the generated summary. 

We use longformer encoder-decoder \cite{beltagy2020longformer} as our summarization model to account for longer documents. We use 4 and 8 exemplars for SuperSCOTUS and CivilSum respectively. Implementation details can be found in App. \ref{impl}.

\subsection{Results}
We use the following models for comparison.  The \emph{w/o exemplars} model serves as the baseline, trained without incorporating any exemplars. All other approaches involve exemplar selection. The \emph{BM25} model uses exemplars retrieved based on BM25 semantic similarity. The \emph{BM25 + DPP (BM25)} model implements a two-stage exemplar selection process where diversity is introduced in the second stage using DPP, but both quality and similarity scores are computed with BM25 scores. Our proposed method, ReLexED (BM25 + DPP (IF)), uses influence function scores to compute quality and similarity during the DPP-based selection process.

\input{text/tab_main_results}

\input{text/tab_analysis}

From Table \ref{tab:main}, we observe that incorporating exemplars significantly improves performance across both datasets, particularly in the UniEval metrics of coherence and fluency. Compared to similarity-based (BM25) exemplar selection, we find that introducing diversity through DPP yields additional gains, underscoring the redundancy of using overly similar examples and emphasizing the importance of considering inter-relationships among exemplars. Overall, computing similarity and quality scores using influence scores ensures a more diverse and informative set of exemplars. Gradients offer a more nuanced notion of similarity, as similar examples often result in nearly identical gradients. This is something that BM25-based scores in DPP fail to capture, leading to a less effective selection of diverse examples. Consequently, our influence-based approach captures deeper relationships among examples, improving summarization quality. Overall, we observe huge improvements in both style-based and faithfulness metrics, with notable or comparable gains in lexical-based measures.

\noindent \textbf{Balancing similarity-diversity of Selected Exemplars:} We analyse exemplar selection methods by measuring (i) how similar are the summaries of selected exemplars to the reference summary of the query instance (EQ) (ii) how similar are the exemplar summaries among themselves (IE). We report the average cosine similarity  computed across each pair of summaries, utilizing embeddings from the LegalBERT model \cite{chalkidis2020legal}. 

As shown in Tab. \ref{tab:retrieval}, incorporating a diversity criterion with DPP, reduces inter-exemplar similarity (IE) promoting diversity. This also reduces  similarity between the exemplar summaries and the query summary. But as evidenced in Tab. \ref{tab:main}, these  diverse exemplars lead to greater performance improvements compared to the model relying solely on exemplars selected via BM25 indicating that highly similar exemplars tend to focus on the same aspects, leading to redundancy and limiting the  introduction of new information. In contrast, diversifying exemplars allows to capture a broader range of relevant details. While BM25-based score computation in DPP may show term variation, but they often correspond to similar underlying concepts, resulting in the selection of redundant exemplars. Using influence scores (computed through gradients) enables to reliably compute similar examples and applying a diversity criterion suppresses overly similar exemplars, reflected in lower IE scores in Tab. \ref{tab:retrieval}, leading to enhanced performance compared to BM25-based DPP.

\subsection{Case Study}
In \emph{United States Steel Corp. v. Multistate Tax Commission} from SuperSCOTUS dataset, the case presented diverse legal aspects, including compliance with the Compact Clause, challenges under the Commerce Clause, taxpayer rights under the Fourteenth Amendment, and dissenting concerns regarding federal oversight. However, BM25-based exemplar selection focused narrowly on the Compact Clause, retrieving examples that emphasized the Compact's non-interference with federal supremacy, while neglecting other key aspects, such as burdens on interstate commerce and dissenting perspectives. This narrow focus led to summaries that lacked nuance and comprehensive coverage. By contrast, our RELexED framework addressed this limitation through a two-stage exemplar selection strategy that balances relevance and diversity. RELexED retrieved exemplars covering a wider range of legal dimensions, such as the operational impact of the Compact, constitutional concerns, and dissenting opinions. This diversity enriched the summarization process, resulting in summaries that are more coherent, well-rounded, and reflective of the multifaceted nature of the case.

%% file: text/tab_main_results.tex
\begin{table*}[!h]
\centering
\begin{tabular}{|l|c|c|c|c|c|c|c|} \hline
{}                  & \textbf{R-1}   & \textbf{R-2}   & \textbf{R-L}   & \textbf{BS} & \textbf{AS} & \textbf{Coh}  & \textbf{Flu}  \\ \hline
\multicolumn{8}{|c|}{\textbf{SuperSCOTUS}}              \\ \hline
{w/o exemplars}     & {51.17}        & {24.62}        & {28.24}        & {64.12}     & {56.12}      & {69.12}       & {66.20}     \\ 
{BM25}              & {51.87}        & {25.18}        & {28.91}        & {65.28}     & {59.74}      & {72.42}       & {69.12}     \\ 
{BM25 + DPP (BM25)} & {53.12}        & {25.86}        & {30.03}        & {65.47}     & {60.16}      & {73.16}       & {71.29}     \\ 
{BM25 + DPP (IF)}   & \textbf{54.04} & \textbf{26.46} & \textbf{30.84} & \textbf{65.84} & \textbf{61.15} & \textbf{74.86} & \textbf{72.26} \\ \hline
\multicolumn{8}{|c|}{\textbf{CivilSum}}                       \\ \hline
{w/o exemplars}     & {39.74}        & {17.29}        & {29.75}        & {57.33}       & {59.92}       & {63.34}       & {74.40}     \\ 
{BM25}              & {40.66}        & {18.76}        & {30.57}        & {59.13}       & {62.15}       & {65.13}       & {76.21}     \\ 
{BM25 + DPP (BM25)} & {41.18}        & {19.43}        & {31.15}        & \textbf{59.82} & {62.72}       & {66.15}       & \textbf{78.12} \\ 
{BM25 + DPP (IF)}   & \textbf{41.68} & \textbf{20.25} & \textbf{32.06} & {59.68}       & \textbf{63.82} & \textbf{66.53} & {77.75}     \\ \hline

\end{tabular}
\caption{Results on SuperSCOTUS and CivilSum. BS, AS, Coh and Flu denote BERTScore, AlignScore, Coherence and  Fluency respectively. Our method \textbf{RELexED} refers to BM25 +  DPP (IF). RElexED achieves statistically significant improvements over the baseline (w/o exemplars) performance, using the Wilcoxon signed-rank test with a 95\% confidence interval.}
\label{tab:main}
\end{table*}

%% file: text/tab_analysis.tex
\begin{table}[]
\begin{tabular}{|l|c|c|c|c|}
\hline
                  & \multicolumn{2}{c|}{\textbf{SuperSCOTUS}} & \multicolumn{2}{c|}{\textbf{CivilSum}}       \\  \hline
                  & \textbf{EQ}            & \textbf{IE}             & \textbf{EQ}                          & \textbf{IE}   \\ \hline
BM25              & 0.77           & 0.81           & 0.75 & 0.86 \\
+ DPP (BM25) & 0.75           & 0.79           & 0.73                        & 0.84 \\
+ DPP (IF)   & 0.75           & 0.74           & 0.72                         & 0.76 \\ \hline
\end{tabular}
\caption{Comparison of exemplar selection approaches. EQ, IE denotes similarity between Exemplars-Query and Inter-Exemplars.}
\label{tab:retrieval}
\end{table}

%% file: text/conclusion.tex
In this work, we propose RELexED, a retrieval-augmented framework for legal summarization that introduces a two-stage exemplar selection strategy, accounting for the inter-relationships between exemplars. We leverage  determinantal point process to balance relevance and diversity among the retrieved examples, with scores computed using model gradient-based influence functions. Our experiments on two legal datasets demonstrate that RELexED outperforms traditional similarity-based exemplar selection, paving the way for designing more effective methods to incorporate informative exemplars into legal summarization tasks.

%% file: text/limitations.tex
Our experiments are conducted on two specific legal datasets from different jurisdictions. The extent to which our observations hold true for a broader range of legal systems remains an open question. The legal domain is vast and varied, and different jurisdictions may exhibit unique characteristics that impact the generalizability of summarization models and warrant further investigation for understanding cross-jurisdictional adaptation of these models \cite{santosh2024beyond}.

Additionally, our current approach primarily focuses on exemplars based on semantic similarity and semantic diversity. However, it would be beneficial to explore additional features, such as the temporal nature of legal documents \cite{santosh2024chronoslex} and multi-aspect considerations \cite{santosh2024lexabsumm}. Understanding different facets of legal cases, such as procedural history, legal arguments and jurisdictional relevance, could lead to more tailored exemplar selections for each aspect, enhancing the quality of generated summaries.

Our evaluation primarily relies on established summarization metrics which provide a quantitative measure of summarization quality, they may not fully capture the nuanced legal content, context and intricacies essential for legal professionals. This limitation suggests a potential avenue for further research into developing additional legal domain-specific evaluation metrics that can more accurately reflect the complexities inherent in legal documents. Another significant limitation of our study is the absence of direct participation or validation by legal experts in assessing the summarization outputs. We could not conduct expert evaluations due to a lack of access to legal professionals. Engaging legal experts in future evaluations would provide valuable insights into the practical relevance of the summaries.

%% file: text/ethics.tex
All datasets utilized in this study are publicly available and were sourced from legal repositories. While the case documents included in these datasets are not anonymized, we do not anticipate any harm arising from their availability, as the datasets are intended for research purposes and contribute to advancing legal technology.

However, we acknowledge that the datasets may contain inherent biases that reflect the existing legal system, societal norms, and historical case law trends. These biases can manifest in various forms, such as the overrepresentation or underrepresentation of certain legal issues, demographics, or jurisdictions. Consequently, these biases could influence the training process, potentially leading to summaries that reinforce existing disparities within the legal system.  Future work should focus on curating datasets that encompass a wider range of legal contexts, ensuring inclusivity across various jurisdictions and case types. Additionally, ongoing efforts to audit and refine the datasets will be crucial in identifying and correcting biases that may adversely affect the model's performance. . We encourage further research into bias detection and mitigation strategies to enhance the integrity of legal summarization systems.

%% file: text/appendix.tex
\section{DPP Explanation}
\label{dpp_exp}
To understand why $\text{det}(L_k)$ serves as a balanced measure of quality and diversity for a selected set, consider a subset $Y = \{i, j\}$ of elements. The probability of choosing this subset is given as: 
\begin{equation*}
\resizebox{0.7\linewidth}{!}{$
\begin{aligned}
    P(Y ; L) &\propto \det(L_Y) \\
    &= \begin{bmatrix}
    q_i \cdot s_{ii} \cdot q_i & q_j \cdot s_{ij} \cdot q_j \\
    q_j \cdot s_{ji} \cdot q_i & q_j \cdot s_{jj} \cdot q_j
    \end{bmatrix} \\
    &= q_i^2 \cdot q_j^2 \cdot (1- s_{ij}^2)
\end{aligned}
$}
\end{equation*}
If candidate elements are highly relevant, any subset containing them will have a high probability. Conversely, if two candidate elements are similar, any set containing both will have a low probability. Geometrically, this can be interpreted as the squared volume of the space spanned by candidate concept vectors of $Y$, where quality indicates vector length and similarity represents the angle between vectors. This determinant expression turns more complex for larger matrices but a similar intuition holds there. In our case, considering each element as a candidate concept, the final subset achieving the highest probability will include a set of highly relevant concepts while maintaining diversity among them via pairwise repulsion.

\section{Implementation Details}
\label{impl}
We implement our code using Huggingface Transformers library \cite{wolf2020transformers} . We use a learning rate of 1e-4 and select the best model based on the R-1 score on the validation set. The model is trained for 10 epochs with mixed precision, gradient clipping with a maximum norm of 1.0,  early stopping patient set of 3 and optimized end-to-end with Adam optimizer \cite{kingma2014adam}, $beta_1$ = 0.9, and $beta_2$ = 0.999. . We set the maximum query length to 10240/4096; rest of encoder length (16384) is filled with 4/8 exemplars and the maximum output lengths are set to 4096/256 for SuperSCOTUS and CivilSum respectively. We set to filter out top 40 exemplars from first stage selection.